\documentclass[10pt,letterpaper]{article}
\usepackage[top=0.85in,left=2.7in,footskip=0.75in,marginparwidth=2in]{geometry}

\usepackage[utf8]{inputenc}
\usepackage{graphicx}
\usepackage{siunitx}
\usepackage{amsmath}
\usepackage{float}

\usepackage{nameref,hyperref}

\usepackage{microtype}
\DisableLigatures[f]{encoding = *, family = * }

\raggedright
\usepackage{changepage}

\usepackage[aboveskip=1pt,labelfont=bf,labelsep=period,singlelinecheck=off]{caption}

\makeatletter
\renewcommand{\@biblabel}[1]{\quad#1.}
\makeatother

\usepackage{lastpage,fancyhdr,graphicx}
\usepackage{epstopdf}
\fancyheadoffset[L]{2.25in}
\fancyfootoffset[L]{2.25in}

\usepackage{color}

\definecolor{Gray}{gray}{.25}

\usepackage{graphicx}

\usepackage{sidecap}

\usepackage{wrapfig}
\usepackage[pscoord]{eso-pic}
\usepackage[fulladjust]{marginnote}
\reversemarginpar

\begin{document}
\vspace*{0.35in}

\begin{flushleft}
{\Large
\textbf\newline{Three approaches to supervised learning for \\ compositional data with pairwise logratios}
}
\newline
\\
Germà Coenders\textsuperscript{1,*},
Michael Greenacre\textsuperscript{2},
\\
\bigskip
\bf{1} Universitat de Girona, Girona, Spain
\\
\bf{2} Universitat Pompeu Fabra, Barcelona, Spain
\\
\bigskip
* germa.coenders@udg.edu

\end{flushleft}

\section*{Abstract}
The common approach to compositional data analysis is to transform the data by means of logratios. Logratios between pairs of compositional parts (pairwise logratios) are the easiest to interpret in many research problems, and include the well-known additive logratios as particular cases. When the number of parts is large (sometimes even larger than the number of cases), some form of logratio selection is a must, for instance by means of an unsupervised learning method based on a stepwise selection of the pairwise logratios that explain the largest percentage of the logratio variance in the compositional dataset. In this article we present three alternative stepwise supervised learning methods to select the pairwise logratios that best explain a dependent variable in a generalized linear model, each geared for a specific problem. The first method features unrestricted search, where any pairwise logratio can be selected. This method has a complex interpretation if some pairs of parts in the logratios overlap, but it leads to the most accurate predictions. The second method restricts parts to occur only once, which makes the corresponding logratios intuitively interpretable. The third method uses additive logratios, so that $K-1$ selected logratios involve exactly $K$ parts. This method in fact searches for the subcomposition with the highest explanatory power. Once the subcomposition is identified, the researcher's favourite logratio representation may be used in subsequent analyses, not only pairwise logratios. Our methodology allows logratios or non-compositional covariates to be forced into the models based on theoretical knowledge, and various stopping criteria are available based on information measures or statistical significance with the Bonferroni correction. We present an illustration of the three approaches on a dataset from a study predicting Crohn's disease. The first method excels in terms of predictive power, and the other two in interpretability.


\section{Introduction}
Compositional data are data in the form of components, or parts, of a whole, where the relative values of the parts are of interest, not their absolute values.
John Aitchison \cite{Aitchison:82, Aitchison:86, Aitchison:97} pioneered the use of the logratio transformation as a valid, (subcompositionally) coherent way to analyse compositional data.
Coherence means that, if the set of parts is extended or reduced, the relationships between the common parts remain constant, whereas their relative values do change because of the differing sample totals.  
Ratios of parts, however, are invariant with respect to the normalization (closure) of the data and these form the basis of Aitchison's approach to compositional data analysis, often referred to as CoDA.  

Since ratios are themselves compared on a ratio scale, and are usually highly right-skew, they are log-transformed to an interval scale.
Hence, the basic concept and data transformation in CoDA is the \textit{logratio}, with the simplest being the logarithm of a pairwise ratio, for example for two parts $A$ and $B$: $\log(A/B) = \log(A) - \log(B)$. 
This study is concerned with such \textit{pairwise logratios}, denoted henceforth exclusively by the abbreviation LR --- logratios in general will not be abbreviated.
The challenge is to choose a set of LRs that effectively replaces the compositional dataset and are at the same time substantively meaningful to the practitioner as well as having a clear interpretation. 
Once the transformation to LRs is performed, analysis, visualization and inference carries on as before, but always taking into account the interpretation in terms of ratios.

For a composition consisting of $J$ parts, a set of $J-1$ LRs contains the whole information of the composition, as long as each part participates in at least one LR \cite{Greenacre:19}. 
When the number of parts is large (in biological applications often larger than the number of cases), some form of selection of fewer than $J-1$ LRs is convenient or even necessary prior to subsequent statistical analysis.  
Greenacre \cite{Greenacre:18, Greenacre:19} developed an unsupervised learning method based on a \textit{stepwise} selection of the LRs that explained the maximum percentage of logratio variance in the composition itself, where ``explained" is in the linear regression sense. 
In this article we are interested rather in supervised learning, that is, selecting LRs that best explain or predict a target variable.  
Two supervised approaches have already been proposed \cite{Walach:17, Erb:17} to find the LRs which are most related to a qualitative target variable. 
These approaches are bivariate, that is, each chosen LR does not take into account the explanatory power of the remaining chosen LRs. 
In this paper we present three alternative stepwise supervised learning methods to select the LRs that make a net contribution to explaining a dependent variable in a \textit{generalized linear model}, as an alternative to the hybrid approach by \cite{hinton:2021} with much the same aim. 
As opposed to \cite{hinton:2021}, the dependent variable can be of any kind supported by generalized linear models, including binary (Bernouilli), continuous, or count (Poisson). 
The selection method for the LRs is the standard one in stepwise regression with forward selection, geared to deal with three distinct compositional problems. The conceptual simplicity of stepwise regression coupled with that of LRs can be appealing to applied researchers without a sophisticated statistical background, compared to the approach in \cite{hinton:2021}, and is yet flexible enough to accommodate three variants which fit different research objectives. 

In the first variant, any LR is eligible to belong to the model. 
This approach will generally lead to the best predictive power but can be difficult to interpret \cite{Hron:21}. 
It is thus inappropriate when the main or only objective is interpretation. 
In the second variant only LRs whose pairs of parts do not overlap are eligible --- thus if $\log(A/B)$ is selected, $A$ and $B$ are excluded from any other ratios. 
This makes the \textit{log-contrast} vectors of the selected LRs orthogonal and leads to a simpler interpretation as trade-off effects between pairs of parts on the dependent variable.
The third variant aims at selecting a subset of parts (i.e., a subcomposition) with the highest explanatory power. 
Several stopping criteria are possible, optimising information measures or ensuring significance of the logratios with the Bonferroni correction. 
All variants allow the researcher to modify the variables entered at a given step from his or her expert knowledge, as incorporated into the selection process in \cite{Graeve:20, Wood:21, Rey:21}. 

This article adds to the literature on variable selection in explanatory compositions: first, the compositional developments using \textit{regularized regression}, including Lasso and related methods, for example  \cite{Lin:14, Shi:16,  Lu:19, Louzada:20, Combettes:21, Susin:20}; 
second, the unsupervised methods that aim at finding an optimal subcomposition, for example \cite{Hron:13}; 
third, the \textit{discriminative balance} approach \cite{Quinn:20a} identifies ratios between two or three parts in a supervised problem; 
fourth, the \textit{selbal} approach \cite{Rivera:18, Susin:20} selects two subcompositions, one positively related to the dependent variable and one negatively related, and computes the logratio of the geometric mean of the first over the second as predictor (this has been generalised to more than one logratio by \cite{Gordon:21}); 
fifth and finally, additional approaches such as using amalgamations \cite{Quinn:20b, Greenacre:21}, investigator driven search of LRs \cite{Wang:17}, and Bayesian methods \cite{Zhang:21}. 

The article is organised as follows, we first state the problem of stepwise regression in the context of LRs. We next describe the three variants of the algorithm, each geared to solve a specific problem. We next present an application to one of the data sets used by \cite{Rivera:18}. 
The last section concludes with a discussion.

\section{Compositional stepwise regression} 

\subsection{Compositions and their logratios}

A $J$-part composition can be defined as an array of strictly positive numbers for which ratios between them are considered to be relevant \cite{PawlowskyBuccianti:11}: ${\bf x} = \left(x_1,x_2,\ldots, x_J \right)$, with $x_j>0$ for $j=1,2,\ldots, J$. 
Notice that an alternative definition of a composition, which is more realistic in practice, is to define it as consisting of non-negative numbers, thus admitting zeros and using alternative methods that do not rely on ratios, yet approximate logratio methods very closely --- see, for example, \cite{Greenacre:10a, Greenacre:21}.

Logarithms of ratios are more statistically tractable than ratios, and Aitchison \cite{Aitchison:82} presented the first comprehensive treatment of compositions by means of logratios, using the additive logratio transformation (ALR) in which $J-1$ LRs are computed with a common denominator, where the denominator is assumed here, without loss of generality, to be the last part:
\begin{equation}
   \log\left(\frac{x_j}{x_J}\right) = \log(x_j) - \log(x_J) \ , \ j = 1,2,\ldots,J-1, 
\label{alr}
\end{equation}    

This can easily be generalized to any of the possible $J(J-1)/2$ LRs between any two parts \cite{Aitchison:86, Greenacre:18}:
\begin{equation}
   \log\left(\frac{x_j}{x_{j'}}\right) = \log(x_j) - \log(x_{j'}) \ , \ j=1,2,\ldots,j'-1 \ , \ j' = 2,3,\ldots, J . 
\label{lr}
\end{equation}    
The inherent dimensionality of a composition is $J-1$, which means that $J(J-1)/2 - (J-1)$ LRs are redundant and only $J-1$ LRs can participate in a statistical model. 
Greenacre \cite{Greenacre:18, Greenacre:19} showed that taking exactly $J-1$ LRs in such a way that each part participates in at least one LR, always leads to a non-redundant selection. 
But even $J-1$ is too large a number when the composition has many parts, and the aim of this article is to select a subset of fewer LRs that is optimal in some sense.

The most general form of a logratio is the log-contrast, which can be expressed as: 
\begin{equation}
    \begin{bmatrix}
      \alpha_1 & \alpha_2 & \cdots & \alpha_J
    \end{bmatrix}
    \begin{bmatrix}
      \log(x_1) \\
      \log(x_2) \\
      \vdots \\
      \log(x_J)
    \end{bmatrix} = \boldsymbol{\alpha}^{\sf T} \log({\bf x}) \ , {\rm\ where\ } \sum_{j=1}^J \alpha_j = \boldsymbol{\alpha}^{\sf T} {\bf 1} = 0.
\label{logcontrast}
\end{equation}
An LR is a special case with one value in the coefficient vector $\boldsymbol{\alpha}$ equal to $1$ and another equal to $-1$, corresponding to the numerator and denominator parts respectively, and the remaining coefficients equal to zero.

Other ways of computing logratios involving more than two parts have been suggested \cite{Egozcue:03, Egozcue:05, Fiserova:11, Muller:18} with the requirement of orthogonality of the $\boldsymbol{\alpha}$ vectors in the log-contrasts, which has implications for the logratio's interpretation \cite{Hron:21} as shown below. 
Notice that two LRs can also have mutually orthogonal $\boldsymbol{\alpha}$ vectors if they do not overlap, that is, if no part participates in both LRs. 
For instance, in a four-part composition,  the LRs $\log(x_1/x_2)$ and $\log(x_3/x_4)$ have the orthogonal $\boldsymbol{\alpha}$ vectors $[-1,1,0,0]$ and $[0,0,-1,1]$ respectively.
 
 The LRs can be used as dependent, predicted by non-compositional variables \cite{Egozcue:11}, or as explanatory, to predict a non-compositional outcome \cite{AitchisonBaconShone:84}, which should be continuous 
 in a linear regression model. The extension from a linear model to a generalized linear model is straightforward. For instance, if the dependent variable is a count, a \textit{Poisson} regression can be specified, or if the dependent variable is binary, a \textit{logit} model can be specified \cite{Coenders:17}. In this article we are concerned with using LRs as explanatory variables.
 
 \subsection{Stepwise regression}

Logratio selection in linear or generalized linear models belongs to the domain of statistical learning \cite{James:13}, and, more precisely, supervised statistical learning, because the selection is made with the purpose of optimising the explanatory power or the predictive accuracy with respect to an external response variable. 
Stepwise regression is one of the earliest forms of supervised statistical learning and can be adapted both to linear and generalized linear models. 
The forward selection method of stepwise regression is especially interesting due to its ability to handle any number of LRs, even if $J-1$ is larger than the sample size, and is described simply as: 

\begin{itemize}
    \item In the first step, the algorithm selects the LR leading to the lowest residual sum of squares, or its generalization for both linear models and generalized linear models, $-2 \times \log(\textrm{likelihood})$, abbreviated as -2logLik, which is called deviance for some specific models, like the logit model we consider in the application section.
    \item In the second and subsequent steps, the algorithm adds to the equation the LR leading to the strongest reduction in the residual sum of squares, in -2logLik or in the deviance, one LR at a time.
\end{itemize}
Since adding an LR always decreases -2logLik, a stopping criterion is needed in order not to reach the trivial solution with $J-1$ LRs which would imply that no selection has been made. 
This is achieved by means of adding a penalty to -2logLik as a function of the number of selected LRs. 
The many possibilities available to introduce such a penalty makes the stepwise approach very flexible. 
Let $m$ be the number of parameters estimated in the model and $n$ the sample size. 
The most popular penalties are:
\begin{itemize}
    \item the Akaike information criterion (AIC), which minimises ${\rm -2logLik}+ 2m$ ;
    \item the Bayesian information criterion (BIC), which minimises ${\rm -2logLik}+ \log(n) m$ .
\end{itemize}
Notice that $\log(n)>2$ for $n>7$, so that the BIC criterion will generally lead to more parsimonious models with fewer LRs than the AIC criterion. 
For example, if $n=500$ the penalty is $6.215m$.

Another possibility is to set the penalty in such a way that an additional LR is introduced into the model only if it is statistically significant at a given  significance level. 
At first sight this could be achieved if the penalty factor equals the quantile of the $\chi^2$ distribution with 1 degree of freedom and tail area equal to the significance level. 
For example, ensuring that the added LRs are significant at 5\% is equivalent to  a penalty equal to $3.841 m$.

However, this approach is flawed because of multiple testing.
Since $J-1$ non-redundant LRs are simultaneously being tested for inclusion, the significance level has to be defined in more strict terms in order to account for the accumulation of risks arising from multiple testing. 
A popular criterion is the conservative Bonferroni correction which implies using the $\chi^2$ quantile with a tail area equal to the significance level divided by $J-1$. 
For the commonly encountered $J$ and $n$ values, this criterion usually leads to the strongest penalty (and thus to the smallest set of selected LRs and the highest model parsimony). 
For example, if $J-1=10$, the relevant tail area is $0.005$, the $\chi^2$ quantile is $7.879$ and the procedure minimises  ${\rm -2logLik}+7.879 m$.

It is also well-known that estimates and t-values are biased upwards in stepwise regression, because the variables included are those with the highest values for the particular sample \cite{Whittingham:06}. 
This requires independent testing of the final model with a fresh cross-validation sample. 
If the original sample is large enough it is customary to split it randomly, roughly two thirds being used to run the stepwise regression and one third to validate the final model. Even after validation, it must be taken into account that many models may have a similar fit to the data and the procedure has only found one of them \cite{Whittingham:06}, which makes stepwise regression fit for predictive and exploratory purposes, but not for theory testing. 

Having said this, many learning methods for compositional data use stepwise algorithms due to their wide acceptance and conceptual simplicity \cite{Greenacre:18, Greenacre:19, hinton:2021, Hron:13, Rivera:18, Susin:20}, and the existence of several solutions with similar goodness of fit and the need for cross-validation are shared by even the most sophisticated statistical learning methods. 

\subsection{Introducing expert knowledge in stepwise regression}
Expert knowledge can be a crucial complement to data-driven statistical learning for compositional data \cite{Greenacre:19, Wang:17}, serving to overcome the inherent limitations of the stepwise method. 
In this respect, the user should be able to:
\begin{itemize}
    \item force certain theoretically relevant LRs into the regression equation;
    \item force certain theoretically relevant non-compositional covariates into the regression equation;
    \item choose among LRs with approximately the same significance or AIC/BIC improvements.
\end{itemize}
With respect to the last of the above-mentioned options, there are already three published studies \cite{Graeve:20, Wood:21, Rey:21} where the expert with domain knowledge has interacted with the statistical algorithm to make choices of LRs from a list of those competing to enter.
The idea is to present the expert with the ``top 20" LRs, say, in decreasing order of importance in the modelling, that is increasing order of -2logLik.
Those at the top often have very little difference between them statistically, and the expert can agree with the optimal one but could also, at the expense of slightly  worse fit, choose a LR lower down the list which has a preferred interpretation. 
Notice that the third study cited above \cite{Rey:21} does have a supervised learning objective, since the LRs are chosen not to explain total logratio variance, but logratio variance between four groups of samples.

\section{The logratio selection algorithms}
In this section we propose three variants of the forward stepwise selection algorithm for generalized linear models, each geared towards solving a specific compositional problem.

\subsection{Unrestricted search}
The first algorithm is a straightforward adaption of the unsupervised stepwise selection algorithm \cite{Greenacre:19} to a supervised setting. 
Any non-redundant LRs may be selected by the algorithm. 
This implies that if $\log(C/B)$ and $\log(C/A)$ have already been selected, then $\log(B/A)$ is excluded, since $\log(B/A)=\log(C/A)-\log(C/B)$ and would not contribute to explain variance or improve predictions.
More importantly, redundancy implies that many models lead to the same predictions and goodness of fit. 
Selecting $\log(B/A)$ and $\log(C/B)$ leads to the same predictions as $\log(C/B)$ and $\log(C/A)$ and as $\log(B/A)$ and $\log(C/A)$. 
Thus, the stepwise algorithm will provide only one solution chosen at random, but there are possibly more equivalent solutions with the same goodness of fit.
This argument can be extended to any set of LRs with parts forming a cycle in a graph, which indicates redundancy (see \cite{Greenacre:18, Greenacre:19}) --- the chosen LRs have to form an acyclic graph (examples are shown in Fig.~\ref{graphs} in the application later). 
 
The final solution of this algorithm may be a combination of overlapping and non-overlapping pairs of parts. 
For example, supposing there are $J=7$ parts, denoted by $A,B,C,D,E,F,G$. 
The stepwise algorithm might for instance select $\log(B/A)$, $\log(C/B)$ and $\log(G/F)$. 
The pair $G/F$ does not overlap with any other (i.e., the parts $F$ and $G$ participate in only one LR) while the pairs $B/A$ and $C/B$ overlap in part $B$.

The interpretation of models combining overlapping and non-overlapping LRs is all but intuitive \cite{Coenders:20, Hron:21}. 
In the above example with the selection $\log(B/A)$, $\log(C/B)$ and $\log(G/F)$ in the model, the interpretation would be as follows, taking into account that the effects of the explanatory variables have to be interpreted keeping all other variables constant \cite{Coenders:20}. 
The coefficient associated with $\log(G/F)$ is interpreted as increasing $G$  at the expense of decreasing $F$, while keeping the mutual ratios of $A$, $B$ and $C$ constant. 
Since $\log(G/F)$ does not overlap with the remaining LRs, its interpretation is not affected and its coefficient expresses a trade-off between only the numerator and denominator parts. 
Of course, it can be the case that both $G$ and $F$ increase in absolute terms at different rates. However, in relative terms, i.e.~compositionally speaking, there will still be a trade-off.
The coefficient associated with $\log(B/A)$ is interpreted as increasing $B$ at the expense of decreasing A, while keeping constant both $C$ relative to $B$ and $G$ relative to $F$.
Keeping the ratio of $C$ over $B$ constant means that $C$ changes by the same factor as $B$. 
Thus, the coefficient associated to $\log(B/A)$ is interpreted as increasing $B$ and $C$ by a common factor at the expense of decreasing $A$.  
Likewise, the coefficient associated to $\log(C/B)$ is interpreted as increasing $C$ at the expense of decreasing $B$, while keeping the ratios of $B$ over $A$ and $G$ over $F$ constant. 
Keeping the ratio of $B$ over $A$ constant means that $A$ decreases by the same factor as $B$. 
Thus, the coefficient associated with $\log(C/B)$ is interpreted as increasing $C$ while decreasing $A$ and $B$ by a common factor. 
As a result, the effects of overlapping LRs do not correspond to the effects of the trade-offs between the numerator and denominator parts. 
On the one hand, this requires exercising great care in the interpretation task, and on the other it deviates from the objective of choosing LRs that lead to simple interpretation. 

Nevertheless, the present variant of the algorithm selects the LRs that contribute the most to predictive power, overlapping or not. 
If the purpose of the researcher is only to make predictions, then interpretation may not be essential and this variant may be the best choice. The following two variants of the method make for a simpler interpretation.

\subsection{Search for non-overlapping pairwise logratios}
In this variant of the algorithm, the stepwise search is restricted to at most $J/2$ (if $J$ is even) or $(J-1)/2$ (if $J$ is odd) LRs with non-overlapping parts. 
This limitation may yield a lower predictive power in some applications, but may be very welcome for high-dimensional compositions where parsimony is a must. 
Since they are non-overlapping, the $K/2$ selected LRs will involve exactly $K$ parts. 
This approach has some important advantages. 
Non-overlapping LRs have orthogonal $\boldsymbol{\alpha}$ vectors in Eq. \ref{logcontrast} by construction. 
They are thus an exception to the often-quoted problems when interpreting LRs as explanatory variables \cite{Hron:21}. 
For this reason, their effects on the dependent variable can be interpreted in a straightforward manner in terms of trade-offs between only the numerator and the denominator parts \cite{Hron:21}, as intended when building the LRs. 
Each non-overlapping LR can also be considered to be a balance up to a multiplicative scalar, which brings the algorithm close to the discriminative-balance approach in \cite{Quinn:20a}, where ratios between two or three parts are selected in a supervised problem.
This algorithm is faster than the unrestricted one, as it continuously removes LRs from the set of feasible choices.

\subsection{Search for additive logratios in a subcomposition}
This algorithm draws from the fact that a subcomposition with $K$ parts can be fully represented by $K-1$ LRs as long as each part participates in at least one LR \cite{Greenacre:19} and that any logratio selection fulfilling this criterion has identical predictions and goodness of fit \cite{Coenders:20}. 
This includes the additive logratios (ALRs) with any part of the subcomposition in the denominator and the remaining $K-1$ parts in the numerator, which makes for a shorter search of candidate LRs and makes the interpretation easier. 
Thus, this algorithm searches for the $K$-part subcomposition with the highest explanatory power by fixing the denominator part of the LR determined in the first step and then bringing in additional parts as numerators of the entering LRs.

The effects of the selected set of ALRs in the model in the final linear model are not interpretable as trade-offs between pairs of parts \cite{Hron:21} but two alternative simple rules for interpretation are given in \cite{Coenders:20}. 
The first one is to interpret the effects as those of increasing the part in the numerator while decreasing all other parts in the subcomposition by a common factor.
The second alternative rule is to reexpress the ALRs as a log-contrast.

In our previous 7-part example, suppose that $\log(B/G)$ is chosen after the initial $\log(A/G)$, for example: 
\begin{equation}
    Y = b_0 + b_1 \log(A/G) + b_2 \log(B/G),
\label{logcontrast2}
\end{equation}
with the equivalent log-contrast:
\begin{equation}
    b_1 \log(A) + b_2 \log(B) - (b_1+b_2) \log(G) .
\label{logcontrast3}
\end{equation}
The coefficient $b_1$ is the effect of increasing $A$ while  decreasing $B$ and $G$ by a common factor, 
the coefficient $b_2$ is the effect of increasing $B$ while decreasing $A$ and $G$ by a common factor, and $(-b_1-b_2)$ shows the effect of increasing $G$ while decreasing $B$ and $A$ by a common factor. 
If we interpret the log-contrast equation as a whole, increasing the parts with positive log-contrast coefficients at the expense of decreasing the parts with negative log-contrast coefficients leads to an increase in the dependent variable.

As stated previously, this algorithm that results in an equation with ALR predictors, also results in identifying a subcomposition.  
If the researcher prefers other parameter interpretations, the resulting subcomposition can be fitted into the regression model in a subsequent step using the researcher's favourite type of logratio transformation, including those with orthogonal $\boldsymbol{\alpha}$ vectors in Eq. \ref{logcontrast}.  

This algorithm has similar objectives as the approaches of the regularized-regression family \cite{Lin:14,  Shi:16, Lu:19}, which also aim at selecting a subcomposition to explain the non-compositional dependent variable. 
It is also related to the selbal approach \cite{Rivera:18}, which selects two subcompositions, one positively related to the dependent variable and one negatively related, and computes the logratio of the geometric mean of the first over the second as predictor. 
The selbal algorithm constrains the effects of all parts to be equal within the numerator and denominator sets. 
The selbal and regularized-regression approaches have been fruitfully compared \cite{Susin:20}.

Our approach can also be understood as a supervised equivalent of the algorithm presented by \cite{Hron:13}, which is a backward stepwise procedure searching for the subcomposition containing the highest possible percentage of total logratio variance of the original composition.
Notice the difference between ``containing" and ``explaining" variance --- contained variance is the contribution to the total logratio variance, where the contributions of each part in the composition are summed to get the total, whereas explained variance is in the regression sense, where a part can not only explain its own contribution to the variance but also contributions due to intercorrelations with other parts.   

All three methods will be available in the next release of the package \texttt{easyCODA} \cite{Greenacre:18} in \textsf{R} \cite{R:21}, using function \texttt{STEPR}, 
with options \texttt{method=1} (unrestricted search) \texttt{method=2} (non-overlapping search) and \texttt{method=3} (search for a subcomposition by selecting ALRs). 
The user can specify how many steps the algorithm will proceed, or a stopping criterion can be specified, either AIC, BIC or Bonferroni.
Theoretically relevant LRs or covariates can be forced into the regression equation at step 0. 
The selection can also be made one single step at a time, where the researcher is presented with a list of LRs that are competing to enter the model, from which  either the statistically optimal one is chosen or a slightly less optimal one with a more interesting and justified substantive meaning and interpretation. 

\section{Application}
The three approaches to logratio selection are applied to a data set relating Crohn's disease to the microbiome of a group of patients in a pediatric CD cohort study \cite{Gevers:14, Rivera:18}.
The 662 patients with Crohn’s disease (coded as 1) and the 313 without any symptoms (coded as 0) are analysed. 
The operational taxonomic unit (OTU) table was agglomerated to the genus level, resulting in a matrix with $J=48$ genera and a total sample size $n=662+313=975$.
All the genera but one had some zeros, varying from 0.41\% to 79.38\% and overall the zeros accounted for 28.8\% of the values in the $975\times 48$ table of OTU counts.
As in \cite{Rivera:18}, the zeros were substituted with the geometric Bayesian multiplicative replacement method \cite{MartinFernandez:15}.

Since the dependent variable is binary, the appropriate member of the generalized linear model family is the logit model, with the probability $p$ of Crohn's disease expressed as the logit (log-odds) $\log\left(\frac{p}{1-p}\right)$. Positive regression coefficients would indicate associations w\textbf{}ith a higher incidence of Crohn’s disease. For the particular case of logit models the deviance equals -2logLik.

The same data set has been analysed using the selbal approach \cite{Rivera:18}, which contrasts two subcompositions of genera $S_1$ and $S_2$ in a single variable equal to the log-transformed ratio of the respective geometric means.  
Thus the coefficients of the parts for each subcomposition are the same, resulting in the following log-contrast as a predictor of the incidence of Crohn’s disease, where the positive and negative coefficients apply respectively to the 8 parts of $S_1$ in the numerator and the 4 parts of $S_2$ in the denominator (abbreviations of the genera are used --- see the Appendix for the list of full names):

\smallskip

\ 0.2041 log(Dial) + 0.2041 log(Dore) + 0.2041 log(Lact) 
+ 0.2041 log(Egge) 
\newline \phantom{x}
+ 0.2041 log(Aggr) + 0.2041 log(Adle)
 + 0.2041 log(Stre) + 0.2041 log(Osci) 
 \newline \phantom{x}
 $-$ 0.4082 log(Rose) $-$ 0.4082 log(Clos) $-$ 0.4082 log(Bact) 
$-$ 0.4082 log(Pept)

\smallskip

\noindent

The results for our three approaches, with the stopping criterion set to optimise BIC are in the left panel of Tables \ref{results1} to \ref{results3}. Variables are ordered according to entry in the stepwise algorithm. The function which is being optimised is deviance$\, +\, 6.8824\,m$. Table \ref{results1} shows the unrestricted solution.

\begin{table}[h!]
    
    \centering
    \begin{tabular}{lcccccc}
   
     \ &  \multicolumn{3}{c}{BIC penalty}  & \multicolumn{3}{c}{Bonferroni penalty} \\
	Ratio & Estimate & s.e. & p-value & Estimate & s.e. & p-value \\
	\hline
	Stre/Rose & 0.3059 & 0.0320 & $<0.0001$ & 0.3022 & 0.0315 & $<0.0001$ \\
    Dial/Pept & 0.1378 & 0.0235 & $<0.0001$ & 0.1618 & 0.0218 & $<0.0001$ \\
    Dore/Bact & 0.2436 & 0.0376 & $<0.0001$ & 0.2393 & 0.0372 & $<0.0001$ \\
    Aggr/Prev & 0.1025 & 0.0221 & $<0.0001$ & 0.1008 & 0.0220 & $<0.0001$ \\
    Adle/Lach & 0.1107 & 0.0275 & $<0.0001$ & 0.1158 & 0.0273 & $<0.0001$ \\
    Lact/Stre & 0.1489 & 0.0371 & $<0.0001$ & 0.1482 & 0.0364 & $<0.0001$ \\
    Osci/Clos & 0.1645 & 0.0429 & \ \ \ 0.0001 & 0.1688 & 0.0426 & $<0.0001$ \\
    Sutt/Bilo & 0.0889 & 0.0247 & \ \ \ 0.0003 & 0.0873 & 0.0246 & \ \ \ 0.0004 \\
    Clot/Pept & 0.0712 & 0.0264 & \ \ \ 0.0070 \\
    \hline
    \qquad BIC & 932.03 & \ & \  & 932.55
    \end{tabular}
    \vspace{0.2cm}
    \caption{Estimates of the final model with the first approach (unrestricted stepwise search). Ratios have been inverted, where necessary, to make all coefficients positive.}
    \label{results1}
\end{table} 

There is an overlap of the genus Stre in steps 1 and 6, which will cause complications in the interpretation.
In the second version of the algorithm then, the selected LRs differ from the sixth step onward, with very small increases in the BIC, shown in Table \ref{results2}.

\begin{table}[h!]
    
    \centering
    \begin{tabular}{lcccccc}
   
     \ &  \multicolumn{3}{c}{BIC penalty}  & \multicolumn{3}{c}{Bonferroni penalty} \\
	Ratio & Estimate & s.e. & p-value & Estimate & s.e. & p-value \\
	\hline
	Stre/Rose & 0.2377 & 0.0294 & $<0.0001$ & 0.2444 & 0.0291 & $<0.0001$ \\
    Dial/Pept & 0.1570 & 0.0221 & $<0.0001$ & 0.1702 & 0.0217 & $<0.0001$ \\
    Dore/Bact & 0.2322 & 0.0379 & $<0.0001$ & 0.2272 & 0.0371 & $<0.0001$ \\
    Aggr/Prev & 0.1026 & 0.0223 & $<0.0001$ & 0.1087 & 0.0222 & $<0.0001$ \\
    Adle/Lach & 0.1077 & 0.0279 & \ \ \ 0.0001 & 0.1139 & 0.0276 & $<0.0001$ \\
    Rumi/Clos &	0.2511 & 0.0660	& \ \ \ 0.0001 & 0.2553 & 0.0642 & $<0.0001$ \\
    Sutt/Bilo & 0.0728 & 0.0248 & \ \ \ 0.0033 & 0.0844 & 0.0245 & \ \ \ 0.0006 \\
    Osci/Faec & 0.1220 & 0.0346 & \ \ \ 0.0004 & 0.1088 & 0.0333 & \ \ \ 0.0011 \\
    Lact/Turi & 0.0864 & 0.0295 & \ \ \ 0.0034 \\			
    Egge/Euba &	0.0792 & 0.0303 & \ \ \ 0.0091 \\
    \hline
    \qquad BIC & 937.92 & \ & \ & 939.64		
    \end{tabular}
    \vspace{0.2cm}
    \caption{Estimates of the final model with the second approach (non-overlapping LRs). Ratios have been inverted, where necessary, to make all coefficients positive.}
    \label{results2}
\end{table} 
		
Since non-overlapping LRs have orthogonal $\boldsymbol{\alpha}$ vectors, their interpretation is according to the logratio formulation \cite{Hron:21}. 
That is, the incidence of Crohn's disease is significantly associated with an increase in the relative abundance of each numerator genus at the expense of a decrease in the relative abundance of the respective denominator genus.

The third approach identifies the set of ALRs, shown in Table \ref{results3}, where there is a much bigger increase in the BIC.
The first LR selected, identical to the first ones in the previous results, determines the denominator part of them all.
The algorithm selects 11 ALRs as a 12-part subcomposition.
This subcomposition of 12 genera may be transformed into the practitioner’s favourite logratio representation without changing the predictions or goodness of fit of the model as long as 11 logratios are used \cite{Coenders:20}. 
A convenient example of the former is to rerun the final model with a different part in the ALR denominator. 
This makes it possible to obtain the missing standard error and p-value in the log-contrast corresponding to the denominator part in the original run. 
The remaining estimates and standard errors do not change (Table \ref{rerun}).

If the model is not used merely for prediction, significance of the logratios becomes important. The right panel of Tables \ref{results1} to \ref{rerun} presents the results using a penalty equivalent to forcing the selected LRs to be significant at 0.05 with the Bonferroni inequality: deviance $+ 10.7130\,m$.
This has led to selecting one fewer LR for the first approach and two fewer LRs for the second and third approaches. Given the large sample size, BIC also employs a substantial penalty to the deviance and this is why results are barely affected in this particular application.

\begin{table}[h!]
    \centering
    \begin{tabular}{lcccccc}
     \ &  \multicolumn{3}{c}{BIC penalty}  & \multicolumn{3}{c}{Bonferroni penalty} \\
	Ratio & Estimate & s.e. & p-value & Estimate & s.e. & p-value \\
	\hline
    Stre/Rose & 0.1488 & 0.0444 & \ \ \ 0.0008 & 0.1415 & 0.0438 & \ \ \ 0.0012 \\
    Dial/Rose & 0.1354 & 0.0267 & $<0.0001$ & 0.1407 & 0.0262 & $<0.0001$ \\
    Pept/Rose & -0.1909 & 0.0331 & $<0.0001$ & -0.2065 & 0.0324 & $<0.0001$ \\
    Lact/Rose & 0.1547 & 0.0404 & $<0.0001$ & 0.1420 & 0.0397 & \ \ \ 0.0003 \\
    Bact/Rose & -0.2859 & 0.0521 & $<0.0001$ & -0.2792 & 0.0481 & $<0.0001$ \\
    Dore/Rose & 0.2252 & 0.0483 & $<0.0001$ & 0.2021 & 0.0439 & $<0.0001$ \\
    Adle/Rose & 0.1477 & 0.0375 & $<0.0001$ & 0.1511 & 0.0360 & $<0.0001$ \\
    Aggr/Rose & 0.1381 & 0.0332 & $<0.0001$ & 0.1378 & 0.0328 & $<0.0001$ \\
    Prev/Rose & -0.0905 & 0.0260 & \ \ \ 0.0005 & -0.0920 & 0.0258 & \ \ \ 0.0004 \\
    Osci/Rose & 0.1551 & 0.0439 & \ \ \ 0.0004	\\		
    Clos/Rose & -0.2140 & 0.0723 & \ \ \ 0.0031	\\		
    \hline
    \qquad BIC & 964.44 & \ & \ & 967.42		
    \end{tabular}
    \vspace{0.2cm}
    \caption{Estimates of the final model with the third approach (subcomposition search with ALR). Ratios have been left with the fixed denominator part, hence positive and negative coefficients.}
    \label{results3}
\end{table} 

 The ALR model in Table \ref{results3}, for example in the right panel, is best interpreted when converted into the corresponding log-contrast\cite{Coenders:20}. Coefficients of the numerator parts can be taken directly from the estimates, and the coefficient of the denominator part is the sum of all coefficients with reversed sign and can be taken also from Table \ref{rerun}: 
 
\smallskip

\centerline{$-(0.1415+0.1407-0.2065 \cdots -0.0920) = -0.3375$}

\smallskip

\noindent
Coefficients are arranged in descending order for convenience: 

\smallskip

\ \ 0.2021\,log(Dore)+0.1511\,log(Adle)+0.1420\,log(Lact)+0.1415\,log(Stre) \\
+0.1407\,log(Dial)+0.1378\,log(Aggr)$-$0.0920\,log(Prev)$-$0.2065\,log(Pept) \\ 
$-$0.2792\,log(Bact)$-$0.3375\,log(Rose)

\smallskip

Thus, the likelihood of Crohn’s disease increases with increases in the genera Dore, Adle, Lact, Stre, Dial and Aggr, at the expense of decreases in Rose, Bact, Pept, and Prev.

\begin{figure}[h!] 
\centering
\includegraphics[width=\linewidth]{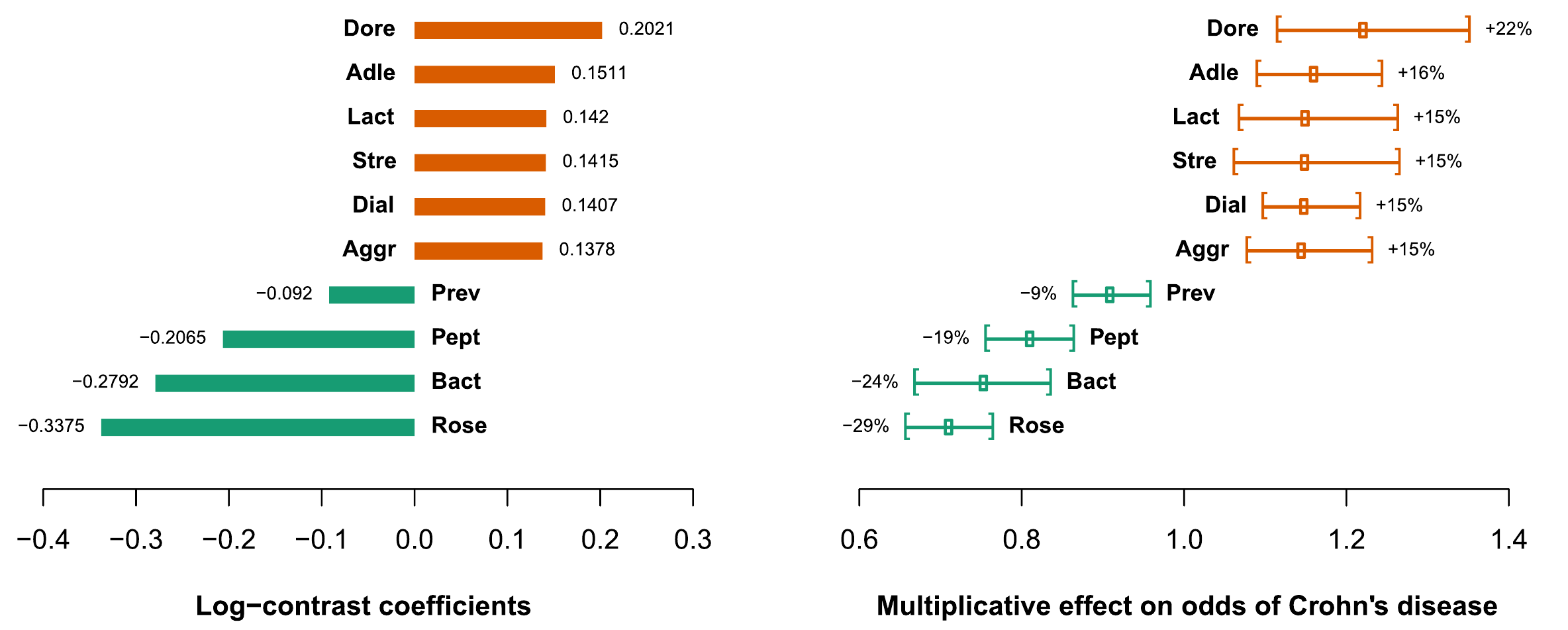}
\vspace{0.1cm}
\caption{Estimated log-contrast coefficients and their conversion to multiplicative effects and 95\% bootstrap confidence intervals.}
\label{effects}
\end{figure}

Fig. \ref{effects} plots the above coefficients as well as the multiplicative effects after exponentiating the coefficients and expressed as percentage effects.
The 95\% bootstrap confidence intervals of these multiplicative effects are shown graphically, based on 1000 bootstrap samples, and the 2.5\% and 97.5\% percentiles of the bootstrapped estimates of the log-contrast coefficients.
It can be seen that none cross the threshold of 1, which represents the hypothesis of no effect for each term of the log-contrast. 

\noindent
As shown above, the coefficients of the parts in the log-contrast are interpreted as the effect on the dependent variable (i.e., the log-odds of having Crohn’s disease) of increasing each part while decreasing all the others by a common factor. Their statistical significance in the form of p-values can be obtained from the estimates of the LR containing the part in the numerator. For instance, the likelihood of Crohn’s disease increases with increases in the genus Dore, at the expense of joint decreases in Adle, Lact, Stre, Dial, Aggr Rose, Bact, Pept, and Prev.

\begin{table}[h!]
    \begin{tabular}{lcccccc}
     \ &  \multicolumn{3}{c}{BIC penalty}  & \multicolumn{3}{c}{Bonferroni penalty} \\
	Ratio & Estimate & s.e. & p-value & Estimate & s.e. & p-value \\
	\hline
    Rose/Stre & -0.3237 & 0.0425 & $<0.0001$ & -0.3375 & 0.0383 & $<0.0001$ \\
    Dial/Stre & 0.1354 & 0.0267 & $<0.0001$ & 0.1407 & 0.0262 & $<0.0001$ \\
    Pept/Stre & -0.1909 & 0.0331 & $<0.0001$ & -0.2065 & 0.0324 & $<0.0001$ \\
    Lact/Stre & 0.1547 & 0.0404 & $<0.0001$ & 0.1420 & 0.0397 & \ \ \ 0.0003 \\
    
     $\vdots$ & $\vdots$ & $\vdots$ & $\vdots$ & $\vdots$ & $\vdots$ & $\vdots$ \\
    Prev/Stre & -0.0905 & 0.0260 & \ \ \ 0.0005 & -0.0920 & 0.0258 & \ \ \ 0.0004 \\
    Osci/Stre & 0.1551 & 0.0439 & \ \ \ 0.0004	\\		
    Clos/Stre & -0.2140 & 0.0723 & \ \ \ 0.0031	\\		
    \hline
    \qquad BIC & 964.44 & \ & \ & 967.42		
    \end{tabular}
    \vspace{0.2cm}
    \caption{Estimates of the final model with the third approach and an alternative denominator (subcomposition search with ALR).}
    \label{rerun}
\end{table} 

In the 9th step in the ALR subcomposition search approach, the 
algorithm reports the Egge genus as the second best choice additional part in the subcomposition after Prev. Since Egge is present in the subcomposition by \cite{Rivera:18}, while Prev is not, a researcher might want to force the logratio with Egge in the numerator instead of Prev as in Table \ref{results3} (Table \ref{expert}). We see that the subcomposition under the BIC penalty (left panel) has changed and the Osci and Clos genera have dropped out and the Egge and Sutt genera have been substituted, at the expense of only a slight increase in the BIC value. In this way, more than one solution can be presented to the user to choose from.

\begin{table}[h!]
    \begin{tabular}{lcccccc}
     \ &  \multicolumn{3}{c}{BIC penalty}  & \multicolumn{3}{c}{Bonferroni penalty} \\
	Ratio & Estimate & s.e. & p-value & Estimate & s.e. & p-value \\
	\hline
    Stre/Rose & 0.1362  &  0.0451 & 0.0025 & 0.1055  &  0.0438  & 0.0161 \\
    Dial/Rose & 0.1283  &  0.0267 & $<0.0001$  &  0.1270  &  0.0263 &  $<0.0001$ \\
    Pept/Rose & -0.2087  &  0.0326 & $<0.0001$  & -0.2045  &  0.0321 &  $<0.0001$ \\
    Lact/Rose &  0.1363  &  0.0399 & 0.0006  &  0.1265  &  0.0390 &  0.0012 \\
    Bact/Rose & -0.3659  &  0.0554 & $<0.0001$  &  -0.3165  &  0.0486 &  $<0.0001$ \\
    Dore/Rose & 0.1847  &  0.0450 & $<0.0001$ &  0.1685  &  0.0445 &  0.0002 \\
    Adle/Rose & 0.1341  &  0.0367 & 0.0003 &  0.1329  &  0.0364 &  0.0003 \\
    Aggr/Rose & 0.1249  &  0.0333 & 0.0002 &  0.1246  &  0.0322 &  0.0001 \\
    Egge/Rose & 0.0823  &  0.0329 & 0.0125 &  0.0905  &  0.0320 &  0.0046  \\
    Prev/Rose & -0.0923  &  0.0268 & 0.0006	\\		
    Sutt/Rose & 0.0964  &  0.0329 & 0.0034	\\		
    \hline
    \qquad BIC & 967.50  & \ & \ & 972.18		
    \end{tabular}
    \vspace{0.2cm}
    \caption{Estimates of the final model with subcomposition search with ALR and Egge/Rose forced into the equation in the 9th step.}
    \label{expert}
\end{table}

\begin{figure}[h!] 
\centering
\includegraphics[width=\linewidth]{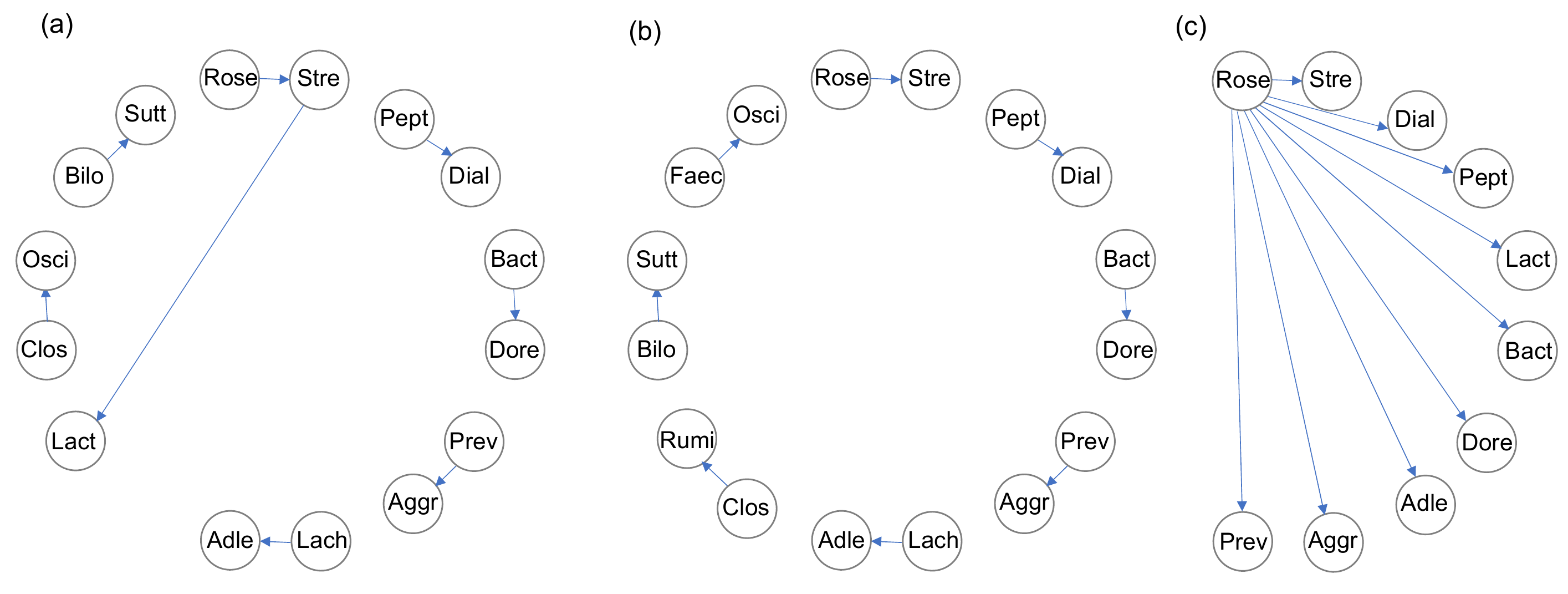}
    \vspace{0.2cm}
\caption{Directed acyclic graphs visualizing the ratios selected in the three stepwise approaches (according to the Bonferroni penalty, the right hand panels in Tables \ref{results1}--\ref{results3}). Arrows point from the denominator to the numerator in every case. In each graph the LR at the top (Stre/Rose) is the first one selected and the following ratios are shown in a clockwise direction. (a) Unrestricted search, showing an overlap of Stre; 15 parts included. (b) Restricted to non-overlap; 16 parts included. (c) ALR selection; 10 parts included, which define a subcomposition, and the only graph out of the three that is connected.}
\label{graphs}
\end{figure}

A convenient way of summarizing the selected LRs of the three algorithms in each of the Tables \ref{results1}--\ref{results3} is in the form of a graph, where the parts are vertices and the LRs defined by the edges \cite{Greenacre:18, Greenacre:19}.
Fig.~\ref{graphs} shows the results according to the Bonferroni penalty in each case.
These are all acyclic graphs, and the ALRs in Fig.~\ref{graphs}(c) define an acyclic connected graph, which is why they explain the LRs in the complete subcomposition \cite{Greenacre:19}.

\begin{figure}[h!] 
\centering
\includegraphics[width=\linewidth]{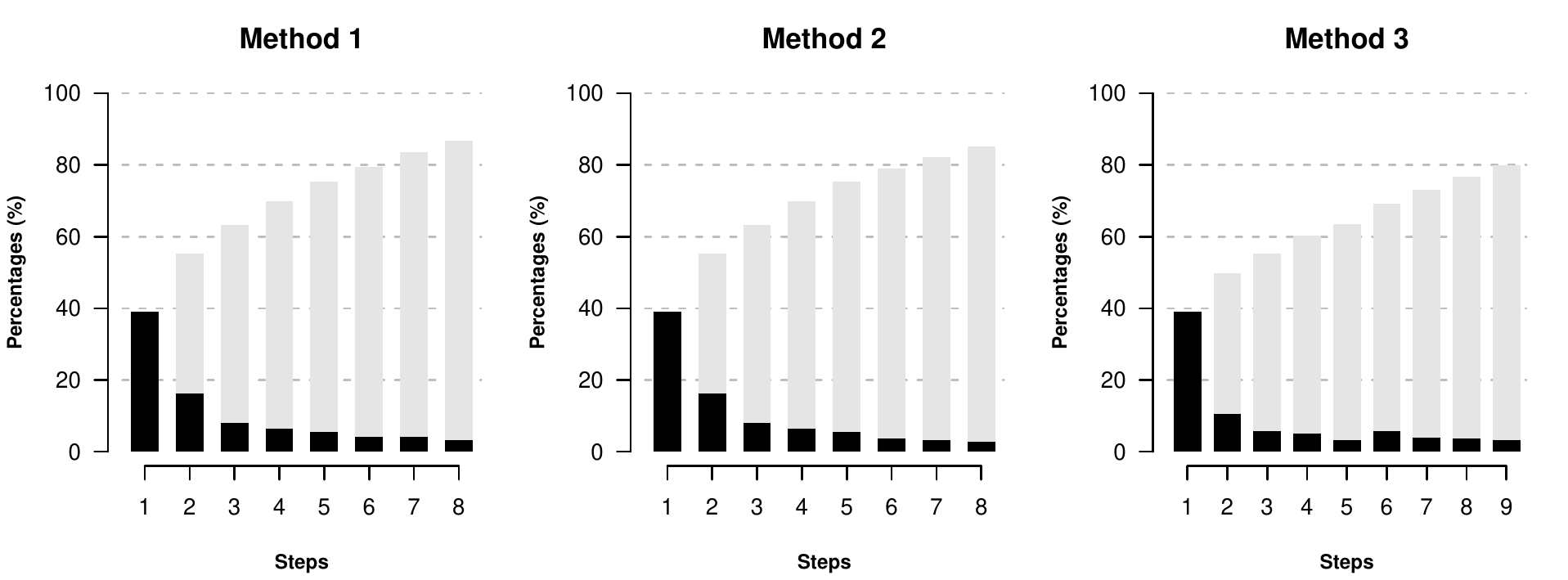}
    \vspace{0.2cm}
\caption{Plots showing incremental amounts (black bars) at each step and cumulative amounts (gray bars) at each step of the three respective algorithms. The values are percentages of the maximum achievable deviance that can be accounted for by using a complete set of $J-1=47$ LRs in the logistic regression.}
\label{scree}
\end{figure}

Fig.~\ref{scree} shows three plots of the sequence of certain diagnostics for the three algorithms.
The null deviance of this application is equal to 1223.9, and if a complete set of $J-1$ logratios is used as predictors, which can be LRs, ALRs or other logratio transformations, the residual (or ``unexplained") deviance is 816.8.
This means that $1223.9 - 816.8 = 407.1$ units of deviance is the best that be accounted for by the LRs.
Using this maximum of 407.1 as 100\%, each LR entering at each step is accounting for a part of that maximum, expressed as a percentage.
In Fig.~\ref{scree} the gray bars show the increasing percentages at each step, which would eventually reach 100\%.
The black bars show the incremental amounts, in a type of scree plot.

\section{Discussion}

The main strength of this article is its conceptual and practical simplicity. Compared to many competing supervised statistical learning methods for compositional data, it yields an actual equation whose predictors the user can actually see, which makes it ultimately possible to introduce modifications based on expert knowledge. The method is very flexible in allowing several types of dependent variables (for the time being, continuous, Poisson and binary), several stopping criteria, and three approaches each geared towards a particular objective, namely prediction, interpretation and subcomposition analysis. The selected LRs are readily interpretable for the latter two modalities. Last but not least, the method will likely  appear familiar to many applied researchers without a sophisticated statistical background, who may gather courage to use and understand it.

The possibility to take advantage of the user's judgement in order to select meaningful albeit statistically suboptimal LRs has already been developed for unsupervised learning  \cite{Greenacre:18, Greenacre:19}. In supervised learning this can also include forcing non-compositional controls into the model and can be a way out of the limitations of purely data-driven approaches \cite{Graeve:20, Wood:21, Rey:21}. The possibility has been shown in the application section by forcing in a part which had been found to be relevant in the previous study by \cite{Rivera:18}. This has led to two alternative subcompositions for the user to choose from, in Table \ref{results3} and Table \ref{expert}.

Among the limitations, the stepwise regression is not a particularly fast algorithm for large numbers of variables, although the third version with ALRs is more scalable than the rest. The results of stepwise regression are indeed sample-dependent and biased. Although this drawback is common to all statistical learning methods, the user has to be reminded that its use is exploratory by nature. 
The way out is to perform cross-validation at every step of the algorithm and then apply the model that is finalized at the last step to a hold-out data set, if one is available. 
Estimates, tests and predictions obtained with the cross-validation sample are unbiased. 
An extension of our approach is then possible to other supervised learning techniques such as classification and regression trees and random forests, where cross-validation is routinely applied.
At each step an LR can be selected to maximize the success of prediction of the response variable, based on cross-validation. 
As we have done in the context of generalized linear models, the stepwise selection can again take place using any of the three selection methods.
The introduction of cross-validation into our approach is the subject of ongoing research. Another future development is to include logratios that involve amalgamated (summed) parts as in \cite{GreenacreGrunskyBaconShone:20, Greenacre:20}.

\section*{Appendix. List of genera abbreviations and full names}

\begin{itemize}
    \itemsep0em 
    \item    Adle: Adlercreutzia
    \item    Aggr: Aggregatibacter
    \item    Bact: Bacteroides
    \item    Bilo: Bilophila
    \item    Clos: Clostridiales
    \item    Clot: Clostridium
    \item    Dial: Dialister
    \item    Dore: Dorea
    \item    Egge: Eggerthella
    \item    Euba: Eubacterium
    \item    Faec: Faecalibacterium
    \item    Lach: Lachnospira
    \item    Lact: Lactobacillales
    \item    Osci: Oscillospira
    \item    Pept: Peptostreptococcaceae
    \item    Prev: Prevotella
    \item    Rose: Roseburia
    \item    Rumi: Ruminococcaceae
    \item    Stre: Streptococcus
    \item    Sutt: Sutterella
    \item    Turi: Turicibacter
\end{itemize}

\section*{Data and software availability}
Microbiome data from 16S rRNA gene sequencing and QIIME 1.7.0 bioinformatics processing were downloaded from Qiita https://qiita.ucsd.edu (study identifier [ID]: 1939).

All methods will be available in the the next release of the \texttt{easyCODA} package \cite{Greenacre:18} in \textsf{R} \cite{R:21}.

\section*{Acknowledgments}
We thank J. Rivera-Pinto, J. J. Egozcue, V. Pawlowsky-Glahn, R. Paredes,  M. Noguera-Julian, and  M. L. Calle for sharing their data with us. 
This work was supported by the Spanish Ministry of Science, Innovation and Universities/FEDER [grant number RTI2018–095518–B–C21]; the Spanish Ministry of Health [grant number CIBERCB06/02/1002]; and the Government of Catalonia [grant number 2017SGR656].

\bibliography{CoendersGreenacre}

\begin{thebibliography}{10}

\bibitem{Aitchison:82}
J.~Aitchison.
\newblock The statistical analysis of compositional data (with discussion).
\newblock {\em J R Stat Soc Ser B}, 44:139--77, 1982.

\bibitem{Aitchison:86}
J.~Aitchison.
\newblock {\em The Statistical Analysis of Compositional Data}.
\newblock Chapman \& Hall, London, 1986.

\bibitem{Aitchison:97}
J.~Aitchison.
\newblock The one-hour course in compositional data analysis, or compositional
  data analysis is simple.
\newblock In V.~Pawlowsky-Glahn, editor, {\em Proceedings of IAMG'97}, pages
  3--35. International Association for Mathematical Geology, 1997.

\bibitem{AitchisonBaconShone:84}
J.~Aitchison and J.~Bacon-Shone.
\newblock Log contrast models for experiments with mixtures.
\newblock {\em Biometrika}, 84:323--330, 1984.

\bibitem{Coenders:17}
G.~Coenders, J.~A. Martín-Fernández, and B.~Ferrer-Rosell.
\newblock When relative and absolute information matter. compositional
  predictor with a total in generalized linear models.
\newblock {\em Stat Model}, 17(6):494--512, 2017.

\bibitem{Coenders:20}
G.~Coenders and V.~Pawlowsky-Glahn.
\newblock On interpretations of tests and effect sizes in regression models
  with a compositional predictor.
\newblock {\em SORT (Stat Oper Res Trans)}, 44:201--220, 2020.

\bibitem{Combettes:21}
P.~L. Combettes and C.~Müller.
\newblock Regression models for compositional data: general log-contrast
  formulations, proximal optimization, and microbiome data applications.
\newblock {\em Stat Biosci}, 13:217--242, 2021.

\bibitem{Egozcue:05}
J.~J. Egozcue and V.~Pawlowsky-Glahn.
\newblock Groups of parts and their balances in compositional data analysis.
\newblock {\em Math Geol}, 37:795--828, 2011.

\bibitem{Egozcue:11}
J.~J. Egozcue, V.~Pawlowsky-Glahn, J.~Daunis-i Estadella, K.~Hron, and
  P.~Filzmoser.
\newblock Simplicial regression. the normal model.
\newblock {\em J Appl Probab Stat}, 6:87--108, 2011.

\bibitem{Egozcue:03}
J.~J. Egozcue, V.~Pawlowsky-Glahn, G.~Mateu-Figueras, and C.~Barceló-Vidal.
\newblock Isometric logratio transformations for compositional data analysis.
\newblock {\em Math Geol}, 35:279--300, 2003.

\bibitem{Erb:17}
I.~Erb, T.~P. Quinn, D.~Lovell, and C.~Notredame.
\newblock Differential proportionality --- a normalization-free approach to
  differential gene expression.
\newblock {\em bioRxiv}, 2017.

\bibitem{Fiserova:11}
E.~Fišerová and K.~Hron.
\newblock On interpretation of orthonormal coordinates for compositional data.
\newblock {\em Math Geosc}, 43(4):455--468, 2011.

\bibitem{Gevers:14}
D.~Gevers, S.~Kugathasan, L.~A. Denson, and et~al.
\newblock The treatment-naïve microbiome in newonset crohn’s disease.
\newblock {\em Cell Host Microbe}, 15:382–392, 2014.

\bibitem{Gordon:21}
E.~Gordon-Rodriguez, T.~P. Quinn, and J.~P. Cunningham.
\newblock Learning sparse log-ratios for high-throughput sequencing data.
\newblock {\em Bioinformatics}, pages e--pub ahead of print, 2021.

\bibitem{Graeve:20}
M.~Graeve and M.~Greenacre.
\newblock The selection and analysis of fatty acid ratios: A new approach for
  the univariate and multivariate analysis of fatty acid trophic markers in
  marine organisms.
\newblock {\em Limnol Oceanogr Methods}, 18:196--210, 2020.

\bibitem{Greenacre:10a}
M.~Greenacre.
\newblock Log-ratio analysis is a limiting case of correspondence analysis.
\newblock {\em Math Geosc}, 42:129--34, 2010.

\bibitem{Greenacre:18}
M.~Greenacre.
\newblock {\em Compositional Data Analysis in Practice}.
\newblock Chapman \& Hall / CRC Press, Boca Raton, Florida, 2018.

\bibitem{Greenacre:19}
M.~Greenacre.
\newblock Variable selection in compositional data analysis using pairwise
  logratios.
\newblock {\em Math Geosc}, 51:649--82, 2019.

\bibitem{Greenacre:20}
M.~Greenacre.
\newblock Amalgamations are valid in compositional data analysis, can be used
  in agglomerative clustering, and their logratios have an inverse
  transformation.
\newblock {\em Appl Comput Geosc}, 5:100017, 2020.

\bibitem{Greenacre:21}
M.~Greenacre.
\newblock Compositional data analysis.
\newblock {\em Annu Rev Stat Appl}, 8:271--99, 2021.

\bibitem{GreenacreGrunskyBaconShone:20}
M.~Greenacre, E.~Grunsky, and J.~Bacon-Shone.
\newblock A comparison of amalgamation and isometric logratios in compositional
  data analysis.
\newblock {\em Comput Geosc}, 148:104621, 2020.

\bibitem{hinton:2021}
A.~Hinton and P.~J. Mucha.
\newblock A simultaneous feature selection and compositional association test
  for detecting sparse associations in high-dimensional metagenomic data.
\newblock {\em Res Square}, 2021.

\bibitem{Hron:21}
K.~Hron, G.~Coenders, P.~Filzmoser, J.~Palarea-Albaladejo, M.~Faměra, and
  T.~M. Grygar.
\newblock Analysing pairwise logratios revisited.
\newblock {\em Comput Geosc}, 53:1643--1666, 2021.

\bibitem{Hron:13}
K.~Hron, P.~Filzmoser, S.~Donevska, and E.~Fišerová.
\newblock Covariance-based variable selection for compositional data.
\newblock {\em Math Geosc}, 45(4):487--488, 2013.

\bibitem{James:13}
G.~James, D.~Witten, T.~Hastie, and R.~Tibshirani.
\newblock {\em An Introduction to Statistical Learning}.
\newblock Springer, New York, 2013.

\bibitem{Lin:14}
W.~Lin, P.~Shi, R.~Feng, and H.~Li.
\newblock Variable selection in regression with compositional covariates.
\newblock {\em Biometrika}, 101:785--797, 2014.

\bibitem{Louzada:20}
F.~Louzada, T.~K.~O. Shimizu, and A.~K. Suzuki.
\newblock The spike-and-slab lasso regression modeling with compositional
  covariates: An application on brazilian children malnutrition data.
\newblock {\em Stat Methods Med Res}, 29:1434--1446, 2020.

\bibitem{Lu:19}
J.~Lu, P.~Shi, and H.~Li.
\newblock Generalized linear models with linear constraints for microbiome
  compositional data.
\newblock {\em Biometrics}, 75:235--244, 2019.

\bibitem{MartinFernandez:15}
J.~A. Martínez-Fernández, K.~Hron, M.~Templ, P.~Filzmoser, and
  J.~Palarea-Albaladejo.
\newblock Bayesian-multiplicative treatment of count zeros in compositional
  data sets.
\newblock {\em Stat Model}, 15:134--158, 2015.

\bibitem{Muller:18}
I.~Müller, K.~Hron, E.~Fišerová, J.~Šmahaj, P.~Cakirpaloglu, and
  J.~Vančáková.
\newblock Interpretation of compositional regression with application to time
  budget analysis.
\newblock {\em Aust J Stat}, 47:3--19, 2018.

\bibitem{PawlowskyBuccianti:11}
V.~Pawlowsky-Glahn and A.~Buccianti.
\newblock {\em Compositional Data Analysis: Theory and Applications}.
\newblock Wiley, New York, 2011.

\bibitem{Quinn:20b}
T.~P. Quinn and I.~Erb.
\newblock Amalgams: data-driven amalgamation for the dimensionality reduction
  of compositional data.
\newblock {\em NAR Genom Bioinf}, 2(4):lqaa076, 2020.

\bibitem{Quinn:20a}
T.~P. Quinn and I.~Erb.
\newblock Interpretable log contrasts for the classification of health
  biomarkers: a new approach to balance selection.
\newblock {\em mSystems}, 5:e00230--19, 2020.

\bibitem{R:21}
{R Development Core Team}.
\newblock {\em R: A Language and Environment for Statistical Computing}.
\newblock R Foundation for Statistical Computing, Vienna, Austria, 2021.
\newblock {ISBN} 3-900051-07-0, http://www.R-project.org.

\bibitem{Rey:21}
F.~Rey, M.~Greenacre, G.~M. Silva~Neto, J.~Bueno-Pardo, M.~R. Domingues, and
  R.~Calado.
\newblock Fatty acid ratio analysis identifies changes in competent
  meroplanktonic larvae sampled over different supply events.
\newblock {\em Mar Environ Res}, 172:accepted for publication, 2021.

\bibitem{Rivera:18}
J.~Rivera-Pinto, J.~J. Egozcue, V.~Pawlowsky-Glahn, R.~Paredes,
  M.~Noguera-Julian, and M.~L. Calle.
\newblock Balances: a new perspective for microbiome analysis.
\newblock {\em mSystems}, 3:e00053–18, 2018.

\bibitem{Shi:16}
P.~Shi, A.~Zhang, and H.~Li.
\newblock Regression analysis for microbiome compositional data.
\newblock {\em Ann Appl Stat}, 10(2), 2020.

\bibitem{Susin:20}
A.~Susin, Y.~Wang, K.~A. Lê~Cao, and M.~L. Calle.
\newblock Variable selection in microbiome compositional data analysis.
\newblock {\em NAR Genom Bioinf}, 2(2):lqaa029, 2020.

\bibitem{Walach:17}
J.~Walach, P.~Filzmoser, K.~Hron, B.~Walczak, and L.~Najdekr.
\newblock Robust biomarker identification in a two-class problem based on
  pairwise log-ratios.
\newblock {\em Chemometrics Intell Lab Syst}, 171:277--285, 2017.

\bibitem{Wang:17}
T.~Wang and H.~Zhao.
\newblock Structured subcomposition selection in regression and its application
  to microbiome data analysis.
\newblock {\em Ann Appl Stat}, 11:771--791, 2017.

\bibitem{Whittingham:06}
M.~J. Whittingham, P.~A. Stephens, R.~B. Bradbury, and R.~P. Freckleton.
\newblock Why do we still use stepwise modelling in ecology and behaviour?
\newblock {\em J Anim Ecol}, 75(5):1182--1189, 2006.

\bibitem{Wood:21}
J.~Wood and M.~Greenacre.
\newblock Making the most of expert knowledge to analyse archaeological data: A
  case study on parthian and sasanian glazed pottery.
\newblock {\em Archaeol Anthropol Sci}, 13:136, 2021.

\bibitem{Zhang:21}
L.~Zhang, Y.~Shi, R.~R. Jenq, K.-A. Do, and C.~B. Peterson.
\newblock Bayesian compositional regression with structured priors for
  microbiome feature selection.
\newblock {\em Biometrics}, 77(3):824--838, 2021.

\end{thebibliography}
\bibliographystyle{abbrv}

\end{document}